\documentclass{article}
\usepackage{ijcai21}

\usepackage{times}
\usepackage{soul}
\usepackage{url}
\usepackage[hidelinks]{hyperref}
\usepackage[utf8]{inputenc}
\usepackage[small]{caption}
\usepackage{graphicx}
\usepackage{amsmath}
\usepackage{amsthm}
\usepackage{booktabs}
\usepackage{algorithm}
\usepackage{algorithmic}
\title{DPointNet: A Density-Oriented PointNet for 3D Object Detection in Point Clouds}

\author{
    Anonymous
}

\iftrue
\author{
Jie Li$^{1,2}$
\and
Yu Hu$^{1,2}$
\affiliations
$^1$Research Center for Intelligent Computing Systems \\
Institute of Computing Technology, Chinese Academy of Sciences\\
$^2$University of Chinese Academy of Sciences
\emails
\{lijie2019, huyu\}@ict.ac.cn
}
\fi

\begin{document}

\maketitle

\begin{abstract}
  For current object detectors, the scale of the receptive field of feature extraction operators usually increases layer by layer. Those operators are called scale-oriented operators in this paper, such as the convolution layer in CNN, and the set abstraction layer in PointNet++. The scale-oriented operators are appropriate for 2D images with multi-scale objects, but not natural for 3D point clouds with multi-density but scale-invariant objects. In this paper, we put forward a novel density-oriented PointNet (DPointNet) for 3D object detection in point clouds, in which the density of points increases layer by layer. In experiments for object detection, the DPointNet is applied to PointRCNN, and the results show that the model with the new operator can achieve better performance and higher speed than the baseline PointRCNN, which verify the effectiveness of the proposed DPointNet.
\end{abstract}

\section{Introduction}

\begin{figure}[h]
\centering
\includegraphics[width=0.98\columnwidth]{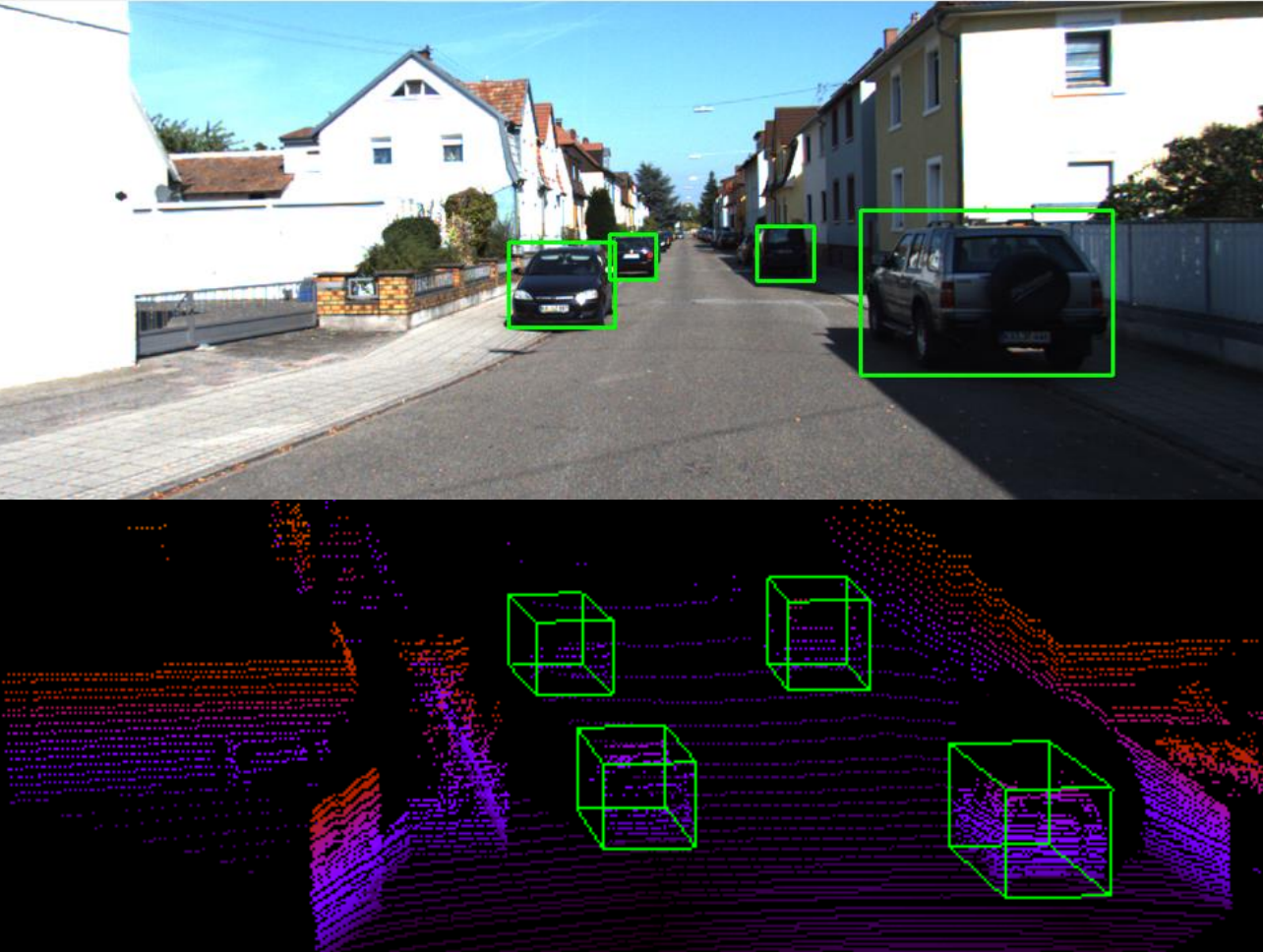}
\caption{Illustration of the density inhomogeneity of point clouds, with the image on the top and the corresponding point cloud on the bottom. Four 3D bounding boxes are plotted in the point cloud for the four cars shown in the image. The car in near areas has much more points than that in far areas.}
\label{fig:problem}
\end{figure}

As one of the key technologies of autonomous driving, 3D object detection can provide accurate spatial location information to help autonomous vehicles make effective predictions and plans. The point clouds from LiDAR are adopted in many 3D detection methods, and the range information is much more accurate and stable than that of camera images. However, some attributes of point clouds are not friendly to regular calculations, such as sparsity, irregularity, large amount of data, and inhomogeneity (as illustrated in Figure~\ref{fig:problem}), and in this paper a novel density-oriented operator is proposed to tackle the inhomogeneity of point clouds.

Most leading 3D detection methods can generally be divided into two categories~\cite{shi2020pv}: grid-based methods, and point-based methods. In grid-based methods, sparse irregular point clouds are transformed into compact regular representations, such as 2D bird-eye-view (BEV) images or 3D voxels. Generally, BEV features are defined manually, such as occupancy, height or reflectance, which are used in PIXOR~\cite{yang2018pixor}. Voxel features are also manually defined in early methods, but after VoxelNet~\cite{zhou2018voxelnet}, Voxel Feature Encoding (VFE) layers are usually adopted to learn features, such as SECOND~\cite{yan2018second}, PointPillars~\cite{lang2019pointpillars}. Unlike grid-based methods that need transformation, point-based methods can directly process the raw point cloud data, and PointNet~\cite{charles2017pointnet} or PointNet++~\cite{qi2017pointnet} are often adopted to learn features of points, such as F-PointNet~\cite{qi2018frustum}, PointRCNN~\cite{shi2019pointrcnn}, 3DSSD~\cite{yang20203dssd}. In addition, some recent methods try to combine the grid-based and point-based modules in one framework. In STD~\cite{yang2019std}, the PointNet++ backbone is used to extract point features, and VFE layers are introduced to convert proposal features from sparse expression to compact representation. PV-RCNN~\cite{shi2020pv} integrates both 3D voxel CNN and PointNet-based set abstraction to learn more discriminative point cloud features. In order to improve the localization precision, in SASSD~\cite{he2020structure}, an auxiliary network is designed to convert the convolutional features back to point-level representations.

Compared with other attributes, the density inhomogeneity of point clouds has received relatively little attention in current methods. However, the point density of objects in point clouds and the scale of objects in images are comparable, while the latter plays an important role in 2D object detection. For the same object, the change of of its distance from the sensor (LIDAR or camera), leads to the change of point density in point clouds or the change of scale in images. The convolution operator for images is scaled-oriented, for the scale of its receptive field usually increases layer by layer, and this property is used in FPN~\cite{lin2017feature} for detecting objects at different scales. Accordingly, we propose a density-oriented operator for point clouds, in which the point density increases layer by layer.

\begin{figure}[h]
\centering
\includegraphics[width=0.98\columnwidth]{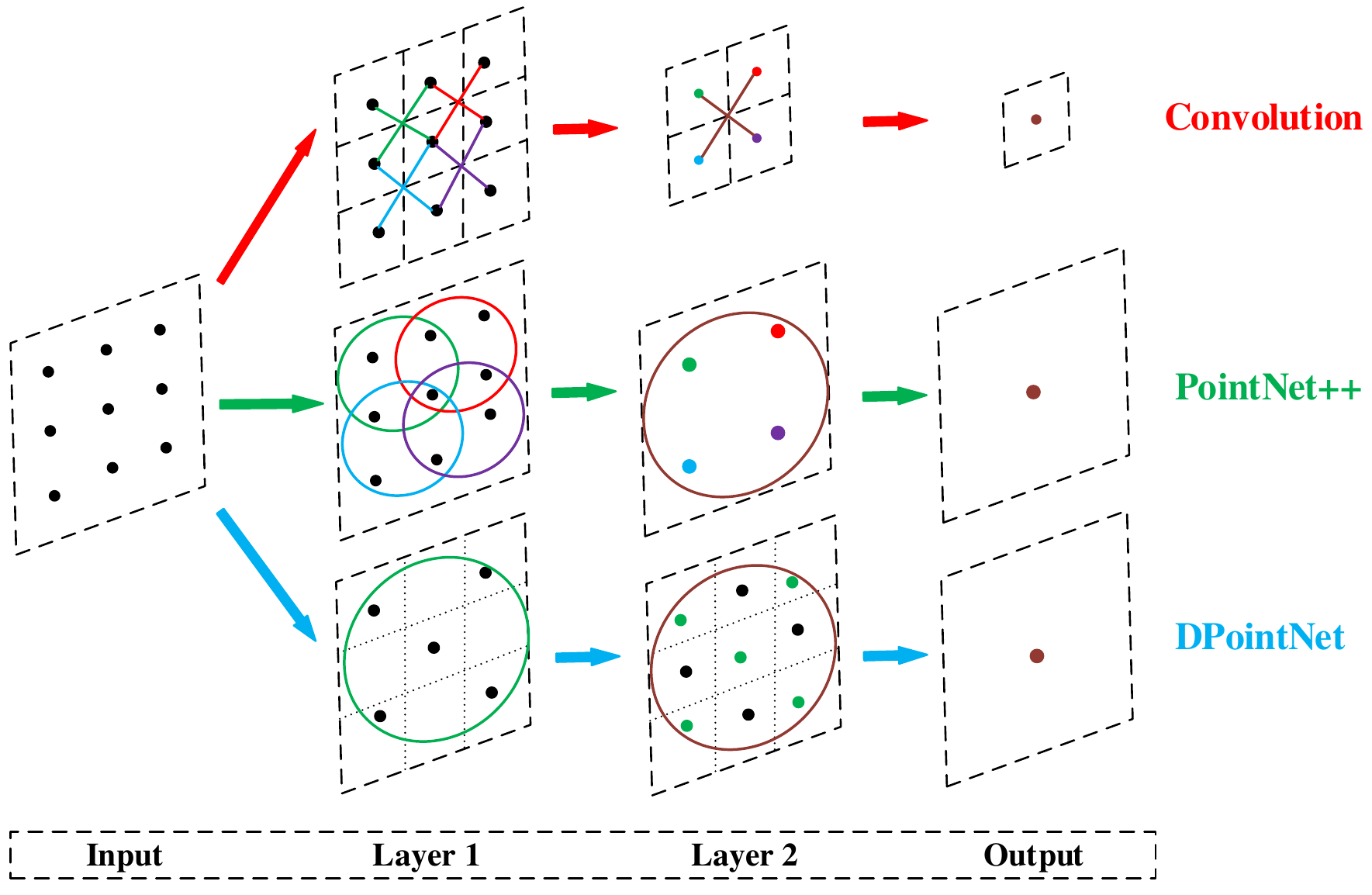}
\caption{Illustration of the calculation processes of three operators, including two scale-oriented operator (convolution and PointNet++) and one density-oriented operator (our DPointNet). Note that `PointNet++' in the figure is used to represent its set abstraction layers.}
\label{fig:operator}
\end{figure}

As shown in Figure~\ref{fig:operator}, the calculation processes of three operators are compared, including two scale-oriented operators (convolution and PointNet++) and one density-oriented operator (our DPointNet). For convolution layers, the local information in the rectangular neighborhood is aggregated through a grid-based calculation, and the scale of the receptive field increases layer by layer. For the set abstraction layers in PointNet++, the local information in the circular (spherical actually) neighborhood is aggregated through a sampling-based calculation, and the receptive field increases with the number of layers.

For the DPointNet, the local information is also aggregated through a sampling-based calculation. As shown in Figure~\ref{fig:operator}, 5 black points are sampled in `Layer 1' and 4 black points are sampled in `Layer 2', and the 5 green points in `Layer 2' represents the fusion information from `Layer 1'. Thus, only part of neighborhood information is considered in each layer, but the latter layer will fuse the information from the previous layer, so the point density in the receptive filed increases layer by layer. In addition, the green circle in `Layer 1' is the same size as the brown circle in `Layer 2', which means the receptive filed of different layers remains unchanged.

Our contributions can be summarized into three-fold.
\begin{itemize}
\item[-] The scale and density attributes of images and point clouds are analyzed, base on which two design principles of density-oriented operators for point clouds are obtained by analogy with the convolution operator for images.

\item[-] A novel density-oriented operator (DPointNet) for point clouds is proposed, which consists of one SG (Sampling and Grouping) layer and several FA (Fusion and Abstraction) layers. To the best of our knowledge, this is also the first density-oriented operator for point clouds.

\item[-] The DPointNet is applied to PointRCNN to verify its effectiveness, and experiments on KITTI dataset show that it has better performance and faster speed than the original PointRCNN with PointNet++.
\end{itemize}

\section{Related Work}

The density inhomogeneity of point clouds has been considered in several methods. In PointNet++, a multi-scale grouping (MSG) strategy for the set abstraction layer is proposed to extract multiple scales of features and combine them to enhance the robustness. In Frustum ConvNet~\cite{wang2019frustum}, a series of frustums at different distances are used to group local points. In Voxel-FPN~\cite{kuang2020voxel}, a multi-resolution voxelization is performed on the point cloud, and a FPN~\cite{lin2017feature} structure is adopted to fuse multi-resolution features. RT3D~\cite{zeng2018rt3d} considers various amounts of valid points in different parts of the car. The method~\cite{wang2019range} uses an adversarial global adaptation and a fine-grained local adaptation, to make the features of far-range objects similar to that of near-range objects. In article~\cite{engels20203d}, two separate detectors are trained to extract features of close-range and long-range objects. SegVoxelNet~\cite{yi2020segvoxelnet} designs a depth-aware head with convolution layers of different kernel sizes and dilated rates, to explicitly model the distribution differences of three parts. In DA-PointRCNN~\cite{li2020a}, a multi-branch backbone network is adopted to match the non-uniform density of point clouds.

In the above-mentioned methods, the density inhomogeneity of point clouds is generally tackled through macro structures instead of micro operators (only PointNet++ tries to adapt its operator to non-uniform densities), and their feature extraction operators are all scale-oriented, such as 2D/3D convolution and PointNet++. In comparison, we propose a density-oriented PointNet (DPointNet) to exploit the inhomogeneity at the operator level, which fits well with the intrinsic properties of point clouds (scale invariance and density inhomogeneity).

\section{DPointNet}
In this section, the proposed DPointNet is described in detail. We first introduce the design principles for density-oriented operators, and then two kinds of network layers in DPointNet are proposed. Finally, the auxiliary head and training losses of the 3D detector with DPointNet are presented.

\begin{figure}[h]
\centering
\includegraphics[width=0.98\columnwidth]{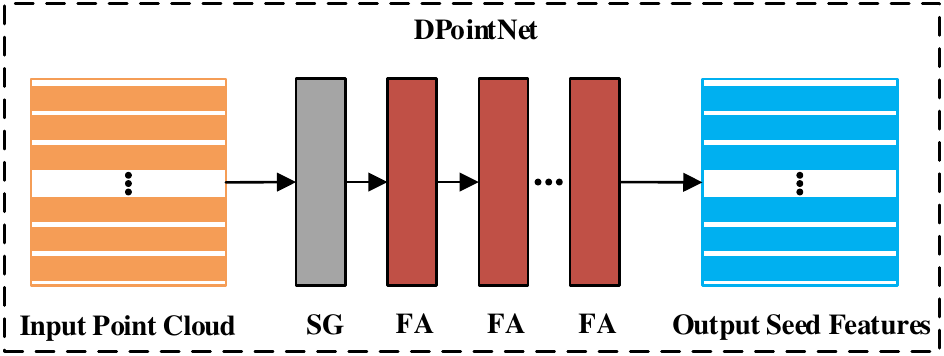}
\caption{An overview of the DPointNet, composed of one SG (Sampling and Grouping) layer and several FA (Fusion and Abstraction) layers.}
\label{fig:dpnet}
\end{figure}

As shown in Figure~\ref{fig:dpnet}, the DPointNet includes one SG (Sampling and Grouping) layer and several FA (Fusion and Abstraction) layers. The SG layer is used to sample seeds and their neighbors, and FA layers are designed to fuse and abstract the features of seeds.

\begin{table*}
\centering
\begin{tabular}{lr|lr}
\toprule
Data  & Characteristics & Operators &  Characteristics\\
\midrule
Images        & scale inhomogeneity   & Scale-oriented   & changing scale \\
              & density invariance    &                  & unchanging density \\
\midrule
Point clouds  & scale invariance      & Density-oriented & unchanging scale \\
              & density inhomogeneity &                  & changing density \\
\bottomrule
\end{tabular}
\caption{Comparison of characteristics for different kinds of data and operators.}
\label{tab:dataop}
\end{table*}

\subsection{Design Principles}
For images and point clouds, scale and density are an interesting pair of attributes. The change of the distance from the same object to sensors, will lead to scale change in images and density change in point clouds. Consequently, two key characteristics are closely related to the design of operators, \emph{i.e.}, scale inhomogeneity and density invariance for images, and scale invariance and density inhomogeneity for point clouds.

As an effective scale-oriented operator, the convolution operator well matches the characteristics of images, with regular grid-based computation (for density invariance) and enlarged receptive filed of multiple layers (for scale inhomogeneity). As shown in Table~\ref{tab:dataop}, the characteristics comparison between images and point clouds, and between scale-oriented and density-oriented operators are presented.

Accordingly, the design principles of density-oriented operators for point clouds can be obtained from the corresponding characteristics:

\begin{itemize}
\item[-]Principle 1: the scale of the receptive field of the operator remains unchanged in different layers.
\item[-]Principle 2: the point density in the receptive field of the operator increases layer by layer.
\end{itemize}


\subsection{Sampling and Grouping Layer}
Given a point cloud, the first step is to sample some seeds and gather the neighbors of seeds, which is done in the SG (Sampling and Grouping) layer of DPointNet.

\begin{figure}[h]
\centering
\includegraphics[width=0.98\columnwidth]{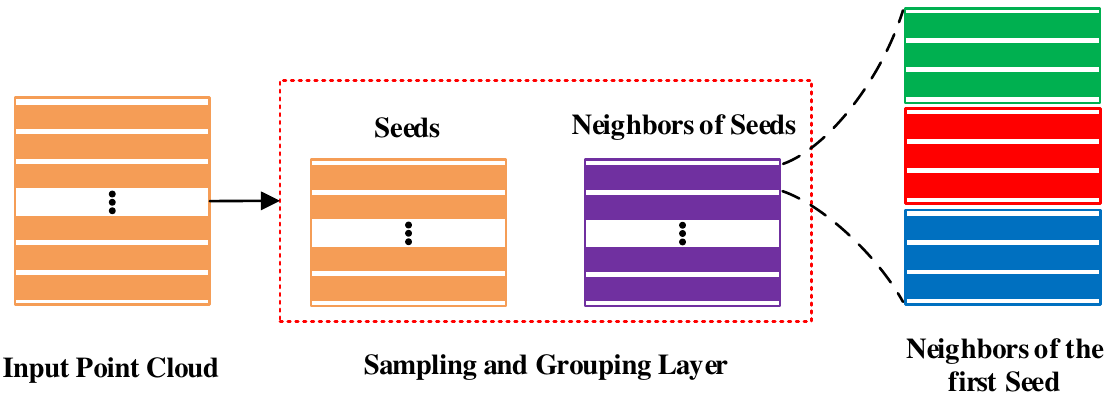}
\caption{Illustration of the Sampling and Grouping Layer in DPointNet.}
\label{fig:sg}
\end{figure}

As shown in Figure~\ref{fig:sg}, the seeds are sampled from the input point cloud by a farthest point sampling (FPS) algorithm, then the ball query is adopted to find the points within a radius to each seed. The number of neighbors of each seed is set to $K$, and repeated random sampling is used if there are not enough neighbors.

The neighbors of each seed are divided into several groups according to the number of operation layers, and the operation layer here refers to the FA (Fusion and Abstraction) layer to be introduced in the next part. For example, three groups are illustrated in Figure~\ref{fig:sg}, \emph{i.e.}, the green group, the red group, and the blue group, which means that the SG layer is followed by three FA layers. All the neighbors used in FA layers are from the SG layer, so there is only one sampling operation, more efficient than PointNet++, in which sampling and grouping are required for each set abstraction layer.

\subsection{Fusion and Abstraction Layer}
\label{sec:fa}
With the neighbor information of seeds, the next step is to fuse and abstract information for each seed, which is done in the FA (Fusion and Abstraction) layers of DPointNet.

\begin{figure}[h]
\centering
\includegraphics[width=0.98\columnwidth]{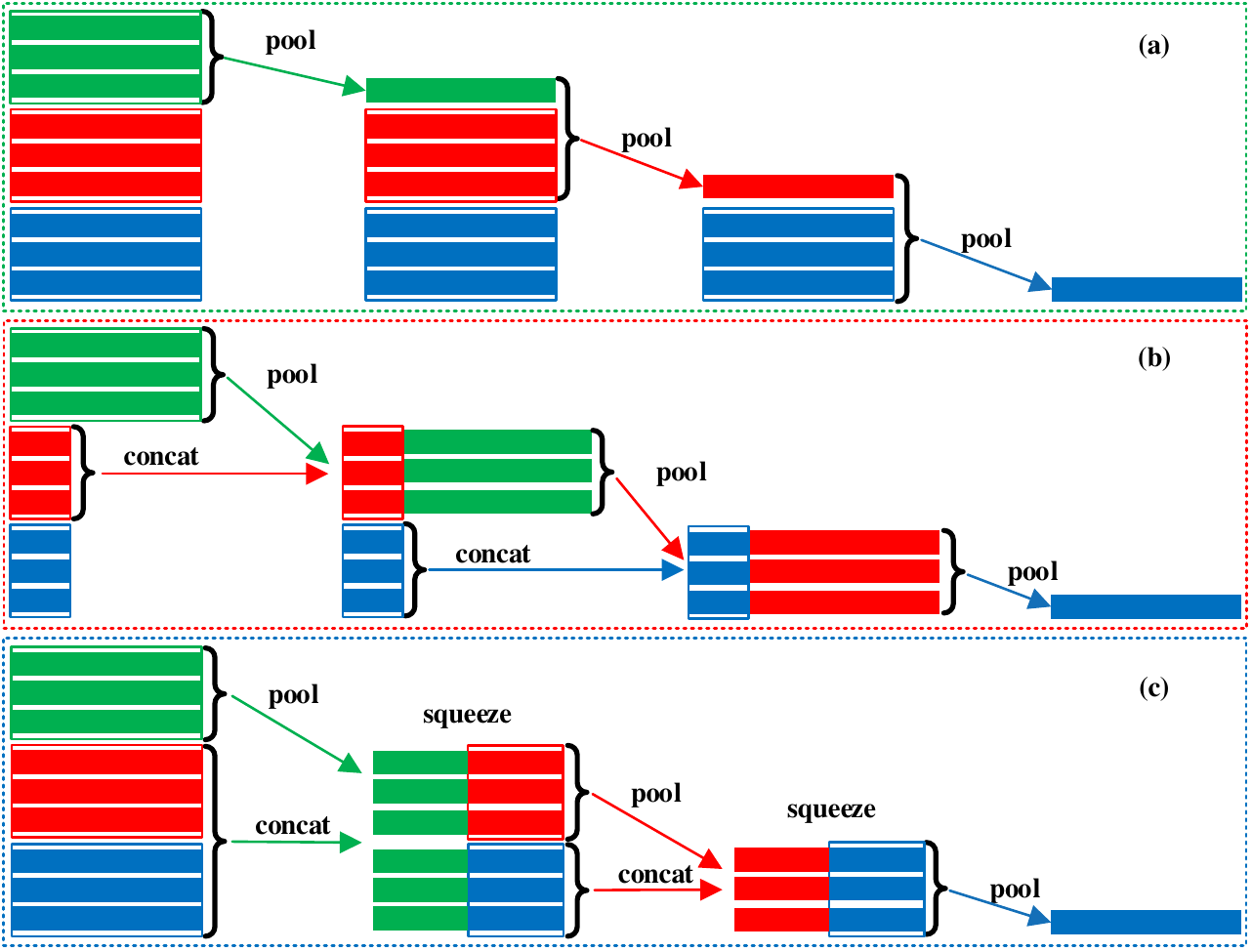}
\caption{Illustration of different schemes to fuse and abstract features: (a) feature appending; (b) coordinate concatenation; (c) feature concatenation. Only the neighbor information of one seed is shown as an example, and the same operations are applied on all the other seeds.}
\label{fig:fa}
\end{figure}

As shown in Figure~\ref{fig:fa}, we design three schemes to fuse and abstract features for the FA layer: (a) feature appending; (b) coordinate concatenation; (c) feature concatenation. Only the calculation of one seed is shown for example, and the same operations are applied on other seeds.

In Figure~\ref{fig:fa}(a), the neighbors of each seed are divided into three groups, and each group is pooled in one FA layer, so the number of groups decreases layer by layer. For each layer, the first group is pooled into one feature, and the feature is appended to the second group, then all the features are transformed by a multi-layer perceptron (MLP) network. Let $g_1$ denote the first group, and $g_{oth}$ denote the other groups, then the calculation of each FA layer is as follows:

\begin{equation}
    mlp\left( {append\left( {pool\left( {{g_1}} \right),{g_{oth}}} \right)} \right)
\label{eq:append}
\end{equation}

In Figure~\ref{fig:fa}(b), the neighbors of each seed are divided into three groups, and each group is processed in one FA layer. For each layer, the first group is features while the other groups are coordinates. The first group is pooled into one feature, and the feature is concatenated with the coordinates of the second group, then an MLP is used to transform the feature. Let $g_1$ denote the first group, and $c_2$ denote the second group, then the calculation is as follows:

\begin{equation}
    mlp\left( {concat\left( {{c_2},pool\left( {{g_1}} \right)} \right)} \right)
\label{eq:coord}
\end{equation}

In Figure~\ref{fig:fa}(c), the neighbors of each seed are divided into three groups, and each group is also processed in one FA layer. For each layer, the first group is pooled into one feature, and the feature is concatenated with the features of other groups, then an MLP is used to squeeze and transform the features.  Let $g_1$ denote the first group, and $g_{oth}$ denote the other groups, then the calculation is as follows:

\begin{equation}
    squeeze\left( {concat\left( {pool\left( {{g_1}} \right),{g_{oth}}} \right)} \right)
\label{eq:concat}
\end{equation}
where \emph{`squeeze'} is also an MLP to reduce the dimensions of the concatenated features.

Scheme (a) can transform the features of all remaining groups in each FA layer, but the feature fusion of the `appending' mechanism is not enough, in which the pooled feature from the first group only affects the pooled feature of the second group. Scheme (b) adopts the `concatenation' mechanism for fusion, and only the coordinate information of the second group is considered to save memory. However, the proportion of the newly added coordinate information is small and decreases sharply layer by layer, which is not suitable for deep network. Scheme (c) combines the advantages of scheme (a) and (b): the sufficient feature extraction from the feature transformation of all remaining groups, and the sufficient feature fusion from the `concatenation' mechanism. An MLP is used to reduce the dimensions of the concatenated features for memory saving. The performance comparison of the three schemes is shown in the experimental section.

\subsection{Auxiliary Head}

\begin{figure}[h]
\centering
\includegraphics[width=0.98\columnwidth]{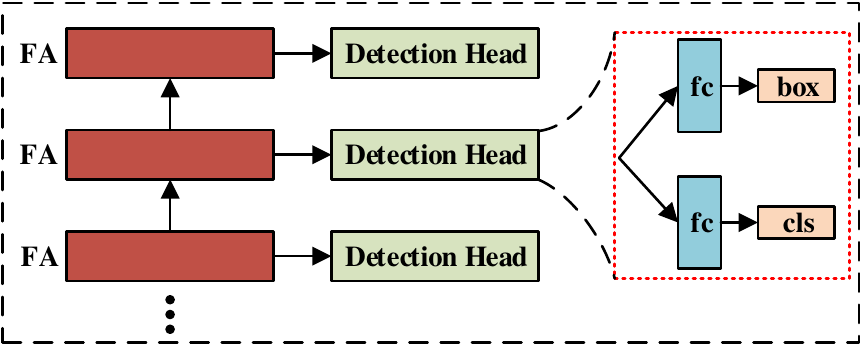}
\caption{Illustration of auxiliary heads for the DPointNet. All detection heads share the same structure, and only the top head is for inference while the others are auxiliary heads for training.}
\label{fig:auxiliary}
\end{figure}

Different from the SA (Set Abstraction) layers in PointNet++, the FA layers in DPointNet do not change the number of seeds, thus each FA layer can provide the features of all seeds. Besides the last FA layer, auxiliary heads can also be adopted for other FA layers. As shown in Figure~\ref{fig:auxiliary}, three detection heads of the same structures are assigned to three FA layers, and only the top head is for inference while the others are auxiliary heads for training.

\begin{figure}[h]
\centering
\includegraphics[width=0.98\columnwidth]{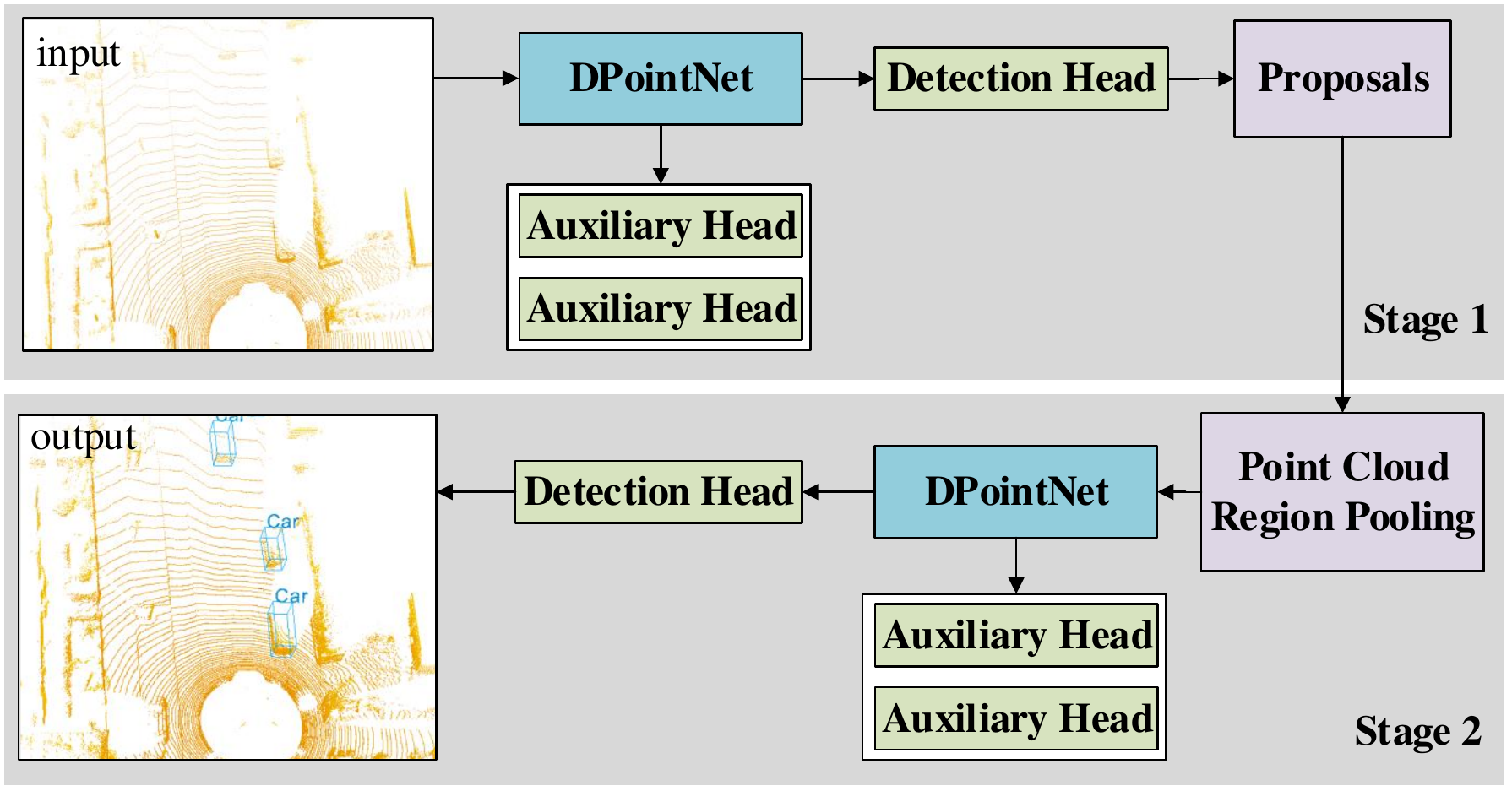}
\caption{The overall architecture of PointRCNN with the proposed DPointNet. The detector consists of two stages: stage 1 for generating 3D proposals, and stage 2 for refining the proposals. The DPointNet is adopted in both stages to learn the features of seeds, and auxiliary heads are added for training process.}
\label{fig:frame}
\end{figure}

As shown in Figure~\ref{fig:frame}, the DPointNet and auxiliary heads are applied to PointRCNN to demonstrate the effectiveness for 3D object detection.

\subsection{Loss Function}
The PointRCNN with the proposed DPointNet (see Figure~\ref{fig:frame}), is trained end-to-end with the region proposal loss $L_{rpn}$ and its auxiliary part $L_{rpn-aux}$, the proposal refinement loss $L_{rcnn}$ and its auxiliary part $L_{rcnn-aux}$, formulated as:
\begin{equation}
    L = {L_{rpn}} + {L_{rpn - aux}} + {L_{rcnn}} + {L_{rcnn - aux}}
\label{eq:loss}
\end{equation}

$L_{rpn}$ and $L_{rpn-aux}$ share the same loss function. The focal loss~\cite{lin2017focal} with default hyper-parameters is adopted to address the foreground-background imbalance, and smooth-L1 loss is utilized for anchor box regression with the predicted residual $\Delta {u^p}$ and the regression target $\Delta {u^t}$:

\begin{equation}
    {L_{rpn}} = {L_{focal}} + \sum\limits_{u \in \left\{ {x,y,z,l,h,w,\theta } \right\}} {{L_{smooth - L1}}\left( {\Delta {u^p},\Delta {u^t}} \right)}
\label{eq:rpnloss}
\end{equation}

$L_{rcnn}$ and $L_{rcnn-aux}$ also share the same loss function. The binary cross entropy loss is used to calculate the classification loss, and and smooth-L1 loss is adopted  for box refinement:

\begin{equation}
    {L_{rcnn}} = {L_{bce}} + \sum\limits_{u \in \left\{ {x,y,z,l,h,w,\theta } \right\}} {{L_{smooth - L1}}\left( {\Delta {u^p},\Delta {u^t}} \right)}
\label{eq:rcnnloss}
\end{equation}

\section{Experiments}
In this section, the PointRCNN with the proposed DPointNet is named as D-PointRCNN for convenience, and the model is evaluated on the widely used KITTI Object Detection Benchmark~\cite{geiger2012are}.

\subsection{Experimental Setup}
\subsubsection{Dataset}
The KITTI dataset contains 7,481 training samples and 7,518 test samples, and the training samples are split into the \emph{train} split (3,712 samples) and the \emph{val} split (3,769 samples). All the models in this paper are trained in the \emph{train} split, and we compare D-PointRCNN and the original PointRCNN on both the \emph{val} split and the \emph{test} split. The dataset includes three categories of Car, Pedestrian and Cyclist, and only the class Car is evaluated for its rich data and scenarios as~\cite{yang20203dssd}.

\subsubsection{Implementation Details}
The D-PointRCNN is developed from the PointRCNN in the OpenPCDet~\cite{openpcdet2020} project, with the DPointNet applied to learn features. For a fair comparison, the main experimental setup remains the same as the original model, and the different parts are introduced in detail.

Data sampling and augmentation strategies remain unchanged: 16,384 points are sampled from each point-cloud, and random flip, scaling, rotation, and GT-AUG~\cite{yan2018second} which adds extra non-overlapping ground-truth boxes from other scenes are employed.

For the network structure, the DPointNet with two auxiliary heads are adopted to replace the original PointNet++. The numbers of FA layers of DPointNet in stage-1 and stage-2 are four and three respectively, the same as the number of SA layers in PointNet++. For memory saving, the number of seeds sampled by the SG layer is 4,096, a quarter of the total sampling points. In addition, the sampling radius and the number of sampling points of each FA layer are 3.0m and 24.

The training scheme is the same as that of PointRCNN in OpenPCDet. The whole D-PointRCNN is trained end-to-end for 80 epochs with the batch size 4 on 1 Tesla V100 GPU, which takes around 20G memory and 18.7 hours (14G memory and 20 hours for PointRCNN), and the learning rate is initialized as 0.01 for Adam optimizer. The final model submitted for testing is trained for 100 epochs and about 23 hours.

\subsection{3D Detection on the KITTI Dataset}

\subsubsection{Evaluation Metric}
The average precision (AP) of class Car with a 0.7 IoU threshold is used to evaluate all results. For the recall positions of AP settings, 40 positions are used for \emph{test} set and 11 positions for \emph{val} split, and all models are trained in \emph{train} split.

\subsubsection{Results on KITTI \emph{test} Set}
As illustrated in Table~\ref{tab:test}, on the main metric, \emph{i.e.} AP on ``moderate'' instances, the D-PointRCNN outperforms PointRCNN by 0.7\%, and similar performance for ``hard'' instances. However, there is about 5\% performance drop for ``easy'' instances, and we think the main cause is the mismatched distribution of \emph{test} and \emph{val} data, as mentioned in Part-$A^2$~\cite{shi2019part} and CIA-SSD~\cite{zheng2020cia}.

\begin{table}[h]
\centering
\smallskip\begin{tabular}{c||c c c }
\toprule
Method & & $A{P_{3D}}(\%)$  \\
 & Easy & Moderate & Hard  \\
\midrule
PointRCNN & \textbf{86.96} & 75.64 & \textbf{70.70}  \\
D-PointRCNN & 81.67 & \textbf{76.34} & 70.34 \\
\bottomrule
\end{tabular}
\caption{Performance comparison on the KITTI \emph{test} set on class Car drawn from official Benchmark. The evaluation metric is Average Precision(AP) with IoU threshold 0.7.}
\label{tab:test}
\end{table}

\subsubsection{Results on KITTI \emph{val} Set}
As illustrated in Table~\ref{tab:val}, the D-PointRCNN outperforms PointRCNN on all three difficulties by 0.4\% to 0.6\%, with only about 60\% running time, demonstrating the effectiveness of the proposed DPointNet. In addition, the memory occupation of the D-PointRCNN ($1.9G$) is similar to that of PointRCNN ($1.7G$). The inference time and memory cost are all tested on 1 Tesla V100 GPU with batch size 1.

\begin{table}[h]
\centering
\smallskip\begin{tabular}{c||c c c|c }
\toprule
Method & & $A{P_{3D}}(\%)$ & &Time \\
 & Easy & Moderate & Hard  & (s) \\
\midrule
*PointRCNN & 88.88 & 78.63 &  77.38 & -  \\
PointRCNN & 88.90 & 78.70 & 77.82  & 0.12 \\
\midrule
D-PointRCNN & \textbf{89.27} & \textbf{79.28} & \textbf{78.35} & \textbf{0.07} \\
\emph{Improvement} & +0.37 & +0.58 & +0.53 & -0.05 \\
\bottomrule
\end{tabular}
\caption{Performance comparison on the KITTI \emph{val} split set on class Car. The PointRCNN with * is the one in the original paper, and the PointRCNN without * is the version in OPenPCDet.}\smallskip
\label{tab:val}
\end{table}

\subsection{Ablation Studies}
In this section, ablation experiments are carried out to analyze the components of the D-PointRCNN, and all models are trained on \emph{train} split and evaluated on \emph{val} split of KITTI dataset.

\subsubsection{The FA Layers of Different Schemes}

As stated in Sec.~\ref{sec:fa}, three schemes are designed for the FA (Fusion and Abstraction) Layer of DPointNet: (a) feature appending; (b) coordinate concatenation; (c) feature concatenation. In terms of design mechanism, scheme (c) has better feature fusion than scheme (a), and better feature abstraction than scheme (b). As shown in Table~\ref{tab:scheme}, scheme (c) outperforms both scheme (a) and (b), demonstrating the effectiveness of the `feature concatenation' mechanism.

\begin{table}[!htbp]

\centering
\smallskip\begin{tabular}{c||c c c}
\toprule
Schemes &  $AP_E$ & $AP_M$ &  $AP_H$    \\
\midrule
scheme (a) & 88.63 & 78.77 & 77.79 \\
scheme (b) & 88.71 & 78.67 & 77.77 \\
scheme (c) & \textbf{89.10} & \textbf{79.01} & \textbf{78.14} \\
\bottomrule
\end{tabular}
\caption{Performance comparison of different schemes for FA layers. $AP_E, AP_M, AP_H$ denote the Average Precision(AP) with IoU threshold 0.7 for easy, moderate, hard difficulty on \emph{val} split.}
\label{tab:scheme}
\end{table}

\subsubsection{The Layer of the Detection Head}
In Figure~\ref{fig:auxiliary}, three detection heads are illustrated, and they are named as ``top head'', ``middle head'', and ``bottom head'' in this part. We conduct experiments to figure out the effects of the layer of the detection head for stage 1.

As shown in Table~\ref{tab:head}, the performance of the ``top head'' is the best among the three layers, which means the deeper features can lead to better results. In D-PointRCNN, only the ``top head'' is used for inference, and the other two are used as auxiliary heads.

In addition, for the stage-1 of D-PointRCNN, only four FA layers are adopted to extract the features of seeds, so the number of FA layers of ``top head'', ``middle head'', and ``bottom head'' are 4, 3, and 2 respectively. As shown in Table~\ref{tab:head}, the largest performance difference is about 0.5\% (for $AP_H$ between ``bottom'' and ``top''), which means a few FA layer can provide relatively good features, thus the computation cost can further be reduced with a small loss of performance.

\begin{table}[!htbp]
\centering
\smallskip\begin{tabular}{c||c c c}
\toprule
Layer &  $AP_E$ & $AP_M$ &  $AP_H$    \\
\midrule
bottom & 88.88 & 78.87 & 77.66 \\
middle & 88.95 & 78.91 & 77.91 \\
top    & \textbf{89.10} & \textbf{79.01} & \textbf{78.14} \\

\bottomrule
\end{tabular}
\caption{Effects of the layer of the detection head.}
\label{tab:head}
\end{table}

\subsubsection{The Sampling Radius of FA Layers}
The receptive fields of all FA layers in DPointNet share the same size, and the effects of different sampling radius, \emph{i.e.}, different sizes of receptive fields, are explored in this part. As shown in Table~\ref{tab:radius}, the optimal value of the radius is around 3.0 m, but the performance difference is marginal for various radius, which reflects the robustness of the operator. In addition, it was found in experiments that the performance variance of 1.0 m or 5.0m is a little larger than that of others, which can be made up by training several times.

\begin{table}[!htbp]
\centering
\smallskip\begin{tabular}{c||c c c}
\toprule
Radius (m) &  $AP_E$ & $AP_M$ &  $AP_H$    \\
\midrule
1.0 & 88.94 & 78.80 & 77.75 \\
2.0 & 88.82 & 78.84 & 77.90 \\
3.0 & \textbf{89.10} & \textbf{79.01} & \textbf{78.14} \\
4.0 & 88.88 & 78.94 & 77.87 \\
5.0 & 88.72 & 78.92 & 77.70 \\
\bottomrule
\end{tabular}
\caption{Effects of different sampling radius.}
\label{tab:radius}
\end{table}

%

\begin{table}[!htbp]
\centering
\smallskip\begin{tabular}{c||c c c|c}
\toprule
Points &  $AP_E$ & $AP_M$ &  $AP_H$    & Memory\\
\midrule

16 & 88.77 & 78.78 & 77.79 & 13~G \\
24 & 89.10 & 79.01 & 78.14 & 18~G \\
32 & \textbf{89.12} & \textbf{79.08} & \textbf{78.16} & 24~G\\

\bottomrule
\end{tabular}
\caption{Effects of different number of sampling points.}
\label{tab:number}
\end{table}

\subsubsection{The Number of Sampling Points of FA Layers}
The number of sampling points of each FA layer is studied in this part, and the results are shown in Table~\ref{tab:number}. More sampling points can provide richer neighborhood information for better performance, but the memory occupation also increases.

Compared with 24 sampling points, 32 points have larger memory cost, but the performance improvement is marginal, so 24 points are adopted in our model to have a good balance between performance and memory cost. In addition, if around 0.5\% performance degradation is allowed, then 16 points could be a good choice to save memory and reduce computation cost.

\begin{table}[!htbp]
\centering
\smallskip\begin{tabular}{c|c|c c c}
\toprule
Auxiliary & Extra &  $AP_E$ & $AP_M$ &  $AP_H$    \\
Head & Training & \\
\midrule
$\times$     & $\times$   &89.10 & 79.01 & 78.14  \\
$\surd$ & $\times$        &89.15 & 79.20 & 78.26  \\
$\surd$ & $\surd$         &\textbf{89.27} & \textbf{79.28} & \textbf{78.35} \\
\bottomrule
\end{tabular}
\caption{Effects of the auxiliary heads and extra training.}
\label{tab:aux}
\end{table}

\subsubsection{Auxiliary Heads and Extra Training}
For the final D-PointRCNN, auxiliary heads and extra training (20 more epochs) are used to improve the performance. As shown in Table~\ref{tab:aux}, the usage of auxiliary heads leads to about 0.2\% improvement on moderate set, and extra training can further bring about 0.1\% performance gain. Auxiliary heads can be removed during the inference stage and extra epochs are only for training, so no additional calculations are introduced for inference.

\section{Conclusion}
We have proposed a density-oriented operator (DPointNet) in this paper, which consists of one SG (Sampling and Grouping) layer and several FA (Fusion and Abstraction) layers. To our knowledge, this is the first density-oriented operator for point clouds. The DPointNet is applied to the PointRCNN to verify its effectiveness on 3D object detection, and it shows better performance and faster speed than the original PointRCNN with PointNet++ operators.

Several parameters of FA layers are studied in experiments, such as sampling radius and number of points, and the results show that the memory occupation and computation cost can be further reduced if a little performance degradation is allowed. In addition, we analyze the scale and density attributes of images and point clouds, and obtain two design principles of density-oriented operators for point cloud, which can provide guidance for the future designs of operators.

\section*{Acknowledgments}
 This work is supported in part by the National Key RD Program of China under grant No. 2018AAA0102701, in part by the Science and Technology on Space Intelligent Control Laboratory under grant No. HTKJ2019KL502003, and in part by the Innovation Project of Institute of Computing Technology, Chinese Academy of Sciences under grant No. 20186090.
\bibliographystyle{named}
\bibliography{ijcai21}

\begin{thebibliography}{}

\bibitem[\protect\citeauthoryear{{Charles} \bgroup \em et al.\egroup
  }{2017}]{charles2017pointnet}
R.~Qi {Charles}, Hao {Su}, Mo~{Kaichun}, and Leonidas~J. {Guibas}.
\newblock Pointnet: Deep learning on point sets for 3d classification and
  segmentation.
\newblock In {\em 2017 IEEE Conference on Computer Vision and Pattern
  Recognition (CVPR)}, pages 77--85, 2017.

\bibitem[\protect\citeauthoryear{{Engels} \bgroup \em et al.\egroup
  }{2020}]{engels20203d}
Guus {Engels}, Nerea {Aranjuelo}, Ignacio {Arganda-Carreras}, Marcos {Nieto},
  and Oihana {Otaegui}.
\newblock 3d object detection from lidar data using distance dependent feature
  extraction.
\newblock In {\em Proceedings of the 6th International Conference on Vehicle
  Technology and Intelligent Transport Systems}, pages 289--300, 2020.

\bibitem[\protect\citeauthoryear{{Geiger} \bgroup \em et al.\egroup
  }{2012}]{geiger2012are}
Andreas {Geiger}, Philip {Lenz}, and Raquel {Urtasun}.
\newblock Are we ready for autonomous driving? the kitti vision benchmark
  suite.
\newblock In {\em 2012 IEEE Conference on Computer Vision and Pattern
  Recognition}, pages 3354--3361, 2012.

\bibitem[\protect\citeauthoryear{{He} \bgroup \em et al.\egroup
  }{2020}]{he2020structure}
Chenhang {He}, Hui {Zeng}, Jianqiang {Huang}, Xian-Sheng {Hua}, and Lei
  {Zhang}.
\newblock Structure aware single-stage 3d object detection from point cloud.
\newblock In {\em CVPR 2020: Computer Vision and Pattern Recognition}, pages
  11873--11882, 2020.

\bibitem[\protect\citeauthoryear{{Kuang} \bgroup \em et al.\egroup
  }{2020}]{kuang2020voxel}
Hongwu {Kuang}, Bei {Wang}, Jianping {An}, Ming {Zhang}, and Zehan {Zhang}.
\newblock Voxel-fpn: Multi-scale voxel feature aggregation for 3d object
  detection from lidar point clouds.
\newblock {\em Sensors}, 20(3):704, 2020.

\bibitem[\protect\citeauthoryear{{Lang} \bgroup \em et al.\egroup
  }{2019}]{lang2019pointpillars}
Alex~H. {Lang}, Sourabh {Vora}, Holger {Caesar}, Lubing {Zhou}, Jiong {Yang},
  and Oscar {Beijbom}.
\newblock Pointpillars: Fast encoders for object detection from point clouds.
\newblock In {\em 2019 IEEE/CVF Conference on Computer Vision and Pattern
  Recognition (CVPR)}, pages 12697--12705, 2019.

\bibitem[\protect\citeauthoryear{{Li} and {Hu}}{2020}]{li2020a}
Jie {Li} and Yu~{Hu}.
\newblock A density-aware pointrcnn for 3d objection detection in point clouds.
\newblock {\em arXiv preprint arXiv:2009.05307}, 2020.

\bibitem[\protect\citeauthoryear{{Lin} \bgroup \em et al.\egroup
  }{2017a}]{lin2017feature}
Tsung-Yi {Lin}, Piotr {Dollar}, Ross {Girshick}, Kaiming {He}, Bharath
  {Hariharan}, and Serge {Belongie}.
\newblock Feature pyramid networks for object detection.
\newblock In {\em 2017 IEEE Conference on Computer Vision and Pattern
  Recognition (CVPR)}, pages 936--944, 2017.

\bibitem[\protect\citeauthoryear{{Lin} \bgroup \em et al.\egroup
  }{2017b}]{lin2017focal}
Tsung-Yi {Lin}, Priya {Goyal}, Ross {Girshick}, Kaiming {He}, and Piotr
  {Dollar}.
\newblock Focal loss for dense object detection.
\newblock In {\em 2017 IEEE International Conference on Computer Vision
  (ICCV)}, pages 2999--3007, 2017.

\bibitem[\protect\citeauthoryear{{Qi} \bgroup \em et al.\egroup
  }{2017}]{qi2017pointnet}
Charles~Ruizhongtai {Qi}, Li~{Yi}, Hao {Su}, and Leonidas~J. {Guibas}.
\newblock Pointnet++: Deep hierarchical feature learning on point sets in a
  metric space.
\newblock In {\em Advances in Neural Information Processing Systems}, pages
  5099--5108, 2017.

\bibitem[\protect\citeauthoryear{{Qi} \bgroup \em et al.\egroup
  }{2018}]{qi2018frustum}
Charles~R. {Qi}, Wei {Liu}, Chenxia {Wu}, Hao {Su}, and Leonidas~J. {Guibas}.
\newblock Frustum pointnets for 3d object detection from rgb-d data.
\newblock In {\em 2018 IEEE/CVF Conference on Computer Vision and Pattern
  Recognition}, pages 918--927, 2018.

\bibitem[\protect\citeauthoryear{{Shi} \bgroup \em et al.\egroup
  }{2019a}]{shi2019pointrcnn}
Shaoshuai {Shi}, Xiaogang {Wang}, and Hongsheng {Li}.
\newblock Pointrcnn: 3d object proposal generation and detection from point
  cloud.
\newblock In {\em 2019 IEEE/CVF Conference on Computer Vision and Pattern
  Recognition (CVPR)}, pages 770--779, 2019.

\bibitem[\protect\citeauthoryear{{Shi} \bgroup \em et al.\egroup
  }{2019b}]{shi2019part}
Shaoshuai {Shi}, Zhe {Wang}, Xiaogang {Wang}, and Hongsheng {Li}.
\newblock Part-$a^{2}$ net: 3d part-aware and aggregation neural network for
  object detection from point cloud.
\newblock 2019.

\bibitem[\protect\citeauthoryear{{Shi} \bgroup \em et al.\egroup
  }{2020}]{shi2020pv}
Shaoshuai {Shi}, Chaoxu {Guo}, Li~{Jiang}, Zhe {Wang}, Jianping {Shi}, Xiaogang
  {Wang}, and Hongsheng {Li}.
\newblock Pv-rcnn: Point-voxel feature set abstraction for 3d object detection.
\newblock In {\em CVPR 2020: Computer Vision and Pattern Recognition}, pages
  10529--10538, 2020.

\bibitem[\protect\citeauthoryear{Team}{2020}]{openpcdet2020}
OpenPCDet~Development Team.
\newblock Openpcdet: An open-source toolbox for 3d object detection from point
  clouds.
\newblock \url{https://github.com/open-mmlab/OpenPCDet}, 2020.

\bibitem[\protect\citeauthoryear{{Wang} and {Jia}}{2019}]{wang2019frustum}
Zhixin {Wang} and Kui {Jia}.
\newblock Frustum convnet: Sliding frustums to aggregate local point-wise
  features for amodal 3d object detection.
\newblock {\em arXiv preprint arXiv:1903.01864}, 2019.

\bibitem[\protect\citeauthoryear{{Wang} \bgroup \em et al.\egroup
  }{2019}]{wang2019range}
Ze~{Wang}, Sihao {Ding}, Ying {Li}, Minming {Zhao}, Sohini {Roychowdhury},
  Andreas {Wallin}, Guillermo {Sapiro}, and Qiang {Qiu}.
\newblock Range adaptation for 3d object detection in lidar.
\newblock In {\em 2019 IEEE/CVF International Conference on Computer Vision
  Workshop (ICCVW)}, pages 2320--2328, 2019.

\bibitem[\protect\citeauthoryear{{Yan} \bgroup \em et al.\egroup
  }{2018}]{yan2018second}
Yan {Yan}, Yuxing {Mao}, and Bo~{Li}.
\newblock Second: Sparsely embedded convolutional detection.
\newblock {\em Sensors}, 18(10):3337, 2018.

\bibitem[\protect\citeauthoryear{{Yang} \bgroup \em et al.\egroup
  }{2018}]{yang2018pixor}
Bin {Yang}, Wenjie {Luo}, and Raquel {Urtasun}.
\newblock Pixor: Real-time 3d object detection from point clouds.
\newblock In {\em 2018 IEEE/CVF Conference on Computer Vision and Pattern
  Recognition}, pages 7652--7660, 2018.

\bibitem[\protect\citeauthoryear{{Yang} \bgroup \em et al.\egroup
  }{2019}]{yang2019std}
Zetong {Yang}, Yanan {Sun}, Shu {Liu}, Xiaoyong {Shen}, and Jiaya {Jia}.
\newblock Std: Sparse-to-dense 3d object detector for point cloud.
\newblock In {\em 2019 IEEE/CVF International Conference on Computer Vision
  (ICCV)}, pages 1951--1960, 2019.

\bibitem[\protect\citeauthoryear{{Yang} \bgroup \em et al.\egroup
  }{2020}]{yang20203dssd}
Zetong {Yang}, Yanan {Sun}, Shu {Liu}, and Jiaya {Jia}.
\newblock 3dssd: Point-based 3d single stage object detector.
\newblock In {\em CVPR 2020: Computer Vision and Pattern Recognition}, pages
  11040--11048, 2020.

\bibitem[\protect\citeauthoryear{{Yi} \bgroup \em et al.\egroup
  }{2020}]{yi2020segvoxelnet}
Hongwei {Yi}, Shaoshuai {Shi}, Mingyu {Ding}, Jiankai {Sun}, Kui {Xu}, Hui
  {Zhou}, Zhe {Wang}, Sheng {Li}, and Guoping {Wang}.
\newblock Segvoxelnet: Exploring semantic context and depth-aware features for
  3d vehicle detection from point cloud.
\newblock {\em arXiv preprint arXiv:2002.05316}, 2020.

\bibitem[\protect\citeauthoryear{{Zeng} \bgroup \em et al.\egroup
  }{2018}]{zeng2018rt3d}
Yiming {Zeng}, Yu~{Hu}, Shice {Liu}, Jing {Ye}, Yinhe {Han}, Xiaowei {Li}, and
  Ninghui {Sun}.
\newblock Rt3d: Real-time 3-d vehicle detection in lidar point cloud for
  autonomous driving.
\newblock {\em IEEE Robotics and Automation Letters}, 3(4):3434--3440, 2018.

\bibitem[\protect\citeauthoryear{Zheng \bgroup \em et al.\egroup
  }{2020}]{zheng2020cia}
Wu~Zheng, Weiliang Tang, Sijin Chen, Li~Jiang, and Chi-Wing Fu.
\newblock Cia-ssd: Confident iou-aware single-stage object detector from point
  cloud.
\newblock {\em arXiv preprint arXiv:2012.03015}, 2020.

\bibitem[\protect\citeauthoryear{{Zhou} and {Tuzel}}{2018}]{zhou2018voxelnet}
Yin {Zhou} and Oncel {Tuzel}.
\newblock Voxelnet: End-to-end learning for point cloud based 3d object
  detection.
\newblock In {\em 2018 IEEE/CVF Conference on Computer Vision and Pattern
  Recognition}, pages 4490--4499, 2018.

\end{thebibliography}

\end{document}